\documentclass[%
 aip,
 amsmath,amssymb,
 reprint,%
]{revtex4-1}

\usepackage{graphicx}
\usepackage{dcolumn}
\usepackage{bm}

\usepackage[utf8]{inputenc}
\usepackage[T1]{fontenc}
\usepackage{mathptmx}
\usepackage{etoolbox}

\makeatletter
\def\@email#1#2{%
 \endgroup
 \patchcmd{\titleblock@produce}
  {\frontmatter@RRAPformat}
  {\frontmatter@RRAPformat{\produce@RRAP{*#1\href{mailto:#2}{#2}}}\frontmatter@RRAPformat}
  {}{}
}%
\makeatother
\begin{document}

\preprint{AIP/123-QED}


\title[Learning unseen coexisting attractors]{Learning unseen coexisting attractors}
\author{Daniel J. Gauthier}
 \affiliation{The Ohio State University, Department of Physics, 191 West Woodruff Ave., Columbus, OH 43210 USA } \affiliation{ResCon Technologies LLC, P.O. Box 21229, Columbus, OH 43221 USA}
  \email{gauthier.51@osu.edu}
\author{Ingo Fischer}%

\affiliation{
Instituto de Física Interdisciplinar y Sistemas Complejos, IFISC (CSIC-UIB), Campus Universitat Illes Balears, E-07122 Palma de Mallorca, Spain
}%

\author{Andr\'e R\"ohm}
\affiliation{Department of Information Physics and Computing, Graduate School of Information Science and Technology, The University of Tokyo, 7-3-1 Hongo, Bunkyo-ku, Tokyo 113-8656, Japan
}%

\date{\today}

\begin{abstract}
Reservoir computing is a machine learning approach that can generate a surrogate model of a dynamical system.  It can learn the underlying dynamical system using fewer trainable parameters and hence smaller training data sets than competing approaches.
Recently, a simpler formulation, known as next-generation reservoir computing, removes many algorithm metaparameters and identifies a well-performing traditional reservoir computer, thus simplifying training even further.  Here, we study a particularly challenging problem of learning a dynamical system that has both disparate time scales and multiple co-existing dynamical states (attractors).
We compare the next-generation and traditional reservoir computer using metrics quantifying the geometry of the ground-truth and forecasted attractors. For the studied four-dimensional system, the next-generation reservoir computing approach uses $\sim 1.7 \times$ less training data, requires $10^3 \times$ shorter `warm up' time, has fewer metaparameters, and has an $\sim 100\times$ higher accuracy in predicting the co-existing attractor characteristics in comparison to a traditional reservoir computer. Furthermore, we demonstrate that it predicts the basin of attraction with high accuracy.
This work lends further support to the superior learning ability of this new machine learning algorithm for dynamical systems. 
\end{abstract}

\maketitle

\begin{quotation}
Reservoir computing is a machine learning approach that is well suited to learning dynamical systems using a short sample of time-series data.  It learns the underlying dynamical system so that the trained model serves as a surrogate or a `digital twin' of the system, and it can reproduce behaviors not contained in the training data set. For chaotic systems, it can reproduce the associated strange attractor and other characteristics.  Surprisingly, the reservoir computer can learn attractors of a dynamical that is has not seen when it is trained on time-series data from a single attractor that it has seen.  Here, we show that a variant machine learning algorithm, known as a next-generation reservoir computer, can reproduce co-existing attractors and the associated basin of attractor using smaller training data sets and with higher accuracy than a traditional reservoir computer.
\end{quotation}

\section{Introduction}\label{sec:Intro}
As human-built dynamical systems become more complex and autonomous, such as advanced propulsion systems or autonomous cars, there is a growing need to develop surrogate models that can be used to rapidly forecast future behaviors, predict boundaries of operation, control the system, or indicate required maintenance, for example.  These surrogate models are often known as `digital twins'.\cite{Grieves2019}  Ideally, they can be traced back to the underlying physics governing the dynamics, be evaluated faster than real time for control applications, and incorporate current observations to improve the model performance.

While many machine learning (ML) algorithms can excel at this task, reservoir computing has emerged as a leading candidate.  A reservoir computer (RC) is an artificial neural network with an input layer through which time-series data is input, a recurrent network (the `reservoir') where each neuron in the network has its own dynamical behavior, and an output layer that is trained for the desired task.  Notably, only the weights of the output layer are trained, often by supervised learning, whereas the weights connecting the input layer to the reservoir and the recurrent connection weights are assigned randomly before training begins and are held fixed thereafter.\cite{Jaeger2004,Maass2002}

The RC excels as a digital twin because of its ability to perform accurate prediction while requiring smaller training data set sizes in comparison to other approaches.\cite{Vlaches2020,Bompas2020}  Also, the model can be trained using only experimental data\cite{Pathak2022} or a hybrid version can be realized that uses both experimental observations and physical model predictions.\cite{Arcomano2022}  The RC paradigm maps well onto physical devices, which can perform the equivalent analog computations to reduce power consumption or increase prediction speed.\cite{Nakajima2021}

Recently, a new ML algorithm was introduced, known as next-generation reservoir computing (NG-RC), that is mathematically equivalent to a traditional RC and shares its attributes\cite{Gauthier2021}, but is much simpler.  It has fewer trainable parameters and hence requires less training data, and it has fewer metaparameters that need to be optimized to obtain high performance in comparison to a traditional RC.  It uses a linear core that consists of a time-delay buffer so that the current and past data points are input to the machine.  The nonlinearity of the reservoir is moved to the output layer, which is a sum of nonlinear functionals of the time-delay data.  The high-weighted nonlinear functionals can be traced back to the physical model of the system, thus peeling back the `black box' nature of many ML algorithms.  Often, polynomial functionals work well, which can be motivated by theories of universal approximators of dynamical systems.\cite{Boyd1985}

Here, we use an NG-RC to create a digital twin of a dynamical system with multiple co-existing attractors and, by coincidence, disparate time scales.  We generate synthetic training data by integrating the differential equations describing the system and using initial conditions that direct the system to only one of the attractors for the training data set.  We then use the trained NG-RC model to predict the other co-existing attractors (unseen during training) of the system by changing the initial conditions.

This problem is particularly challenging:  We previously used a traditional RC for the same problem\cite{Rohm2021} and found modest prediction ability after spending substantial computing time to optimize the algorithm metaparameters. The NG-RC algorithm obtains an accurate solution requiring much less effort to optimize the few remaining metaparameters. The purpose of this paper is to report our approach and document the accuracy of the ML algorithm.

In the next section, we describe the dynamical system introduced by Li and Sprott,\cite{Li2014} which is one of the simplest dynamical systems that displays co-existing attractors.  In Sec.~\ref{sec:NG-RC}, we briefly describe the NG-RC and performance metrics and then give our results in Sec.~\ref{sec:results}.  We end in Sec.~\ref{sec:discussion} with a discussion and point to future directions.

\section{A dynamical system with co-existing attractors}\label{sec:LiSprott}

We focus on systems described by $d$ autonomous nonlinear differential equations of the form
\begin{equation}
    \dot{\mathbf{x}} = \mathbf{f}(\mathbf{x}) \label{eq:diffeq}
\end{equation}
with initial condition $\mathbf{x}_0$. Here, $\dot{\mathbf{x}} \equiv d\mathbf{x}/dt$ and $\mathbf{f}$ is referred to as the vector field.

When the time-series data generated by Eq.~\ref{eq:diffeq} is sampled at discrete times $\{t_n, t_{n-1}, t_{n-2}, \ldots\}$ with step $dt=t_{n}-t_{n-1}$, it is appropriate to consider the flow $\mathbf{F}$ of the dynamical system, which is an evolution function that steps it from one discrete point to the next through the relation
\begin{equation}
\mathbf{x_{n+1}} = \mathbf{F}(\mathbf{x}_n). \label{eq:flow}
\end{equation}
Formally, the flow can be found by integrating Eq.~\ref{eq:diffeq} as
\begin{equation}
    \mathbf{F}(\mathbf{x}_n)=\mathbf{x}_n + \int_{t_n}^{t_{n+1}} \mathbf{f}(\mathbf{x}) ds. \label{eq:formalflow}
\end{equation}

We consider the $d=4$ system introduced by Li and Sprott\cite{Li2014} given by
\begin{eqnarray}
\dot{x} &=& -x+y, \label{eq:x} \\  
\dot{y} &=& -xz+u, \label{eq:y} \\ 
\dot{z} &=& xy - a, \label{eq:z} \\ 
\dot{u} &=& -by,  \label{eq:u}
\end{eqnarray}
where $\mathbf{x}=[x,y,z,u]^T,$ $T$ is the transpose, and $a$ and $b$ are parameters.  Despite its simplicity, Eqs.~\ref{eq:x}-\ref{eq:u} show a variety of behaviors including a large separation of time scales, periodic, quasi-periodic, and chaotic dynamics depending on the parameter values, and they do not have any stable or unstable steady states.  The system has a $\pi$-rotation symmetry about the $z$-axis: $x\rightarrow -x$, $y\rightarrow -y$, $z\rightarrow z$, and $u\rightarrow -u$.

\subsection{Methods} \label{Sec:Methods}

Throughout this study, we choose $a=6$ and $b=0.1$ for which the system has three coexisting attractors: a torus (quasi-periodic behavior) and two chaotic attractors. The large difference in $a$ and $b$ gives rise to a large separation of time scales as will be seen in Sec.~\ref{sec:results} below.  For future reference, the Lyaponov exponents for the chaotic attractors are (0.2520,0.,-0.0052,-1.2467).\cite{Li2014} The inverse of the largest positive exponent, equal to 3.97 time units, is known as the Lyapunov time and gives the characteristic exponential growth time constant of small differences in the attractor trajectories. 

We generate synthetic data by numerically integrating Eqs.~\ref{eq:diffeq} using a variable step size 4$^{th}$-order Runge-Kutta method (RK45, SciPy solve\_ivp) with an absolute error tolerance of $10^{-7}$.  We save the data at regular steps with $dt=0.05$ whose value selection is discussed below.

Run times given below are for running Python 3.9, Numpy 1.20.3, SciPy 1.7.1, and Scikit learn 0.24.2 on an Intel i7-12700 processor running at 2.1 GHz.

\section{Next-generation reservoir computing}\label{sec:NG-RC} 

The task of the NG-RC is to learn the flow $\mathbf{F}(\mathbf{x}_n)$ of the dynamical system using discretely sampled training data, here based on synthetic data generated by integrating Eqs.~\ref{eq:x}-\ref{eq:u}.  Learning the flow has the advantage that it may speed up overall prediction time when $dt$ is taken larger than the typical time step needed to accurately solve numerically the differential equations.  This speed up is important for control applications and may find application in speeding up simulations of systems displaying spatio-temporal chaos.\cite{Barbosa2022}

Briefly, the NG-RC expresses the flow as
\begin{equation}
    \textbf{F}(\textbf{x}_n) = \textbf{x}_n + \mathbb{W}_{out}\mathbf{O}_{total,n}(\textbf{x}_n,\textbf{x}_{n-1},\ldots,\textbf{x}_k), \label{eq:flowNG-RC}
\end{equation}
where $\mathbb{W}_{out}$ is the the set of trainable weights of the output layer and $\mathbf{O}_{total,n}$ is a nonlinear feature vector described below that depends on the current state of the system $\textbf{x}_n$ and $(k-1)$ past states.  Equations~\ref{eq:flow} and \ref{eq:flowNG-RC} describes a discrete dynamical system that has memory, which is controlled by $\mathbb{W}_{out}$ and $k$. To be a universal dynamical system approximator,\cite{Boyd1985,Gonon2020} we must take the limit $k \rightarrow \infty$.  However, it is found that the NG-RC can give highly accurate predictions even when $k$ is small, which depends on $dt$.  For example, when $dt$ is much smaller than the shortest characteristic time of the system, $k=1$ is usually large enough and the highly-weighted terms in $\mathbf{O}_{total,n}$ are largely determined by $\mathbf{f}$.  For larger $dt$, other terms increase in importance but the nonlinear terms in $\mathbf{O}$ can be motivated by the vector field.

The feature vector is a concatenation of a constant term $c$, a linear part $\mathbf{O}_{lin}$, and a nonlinear part $\mathbf{O}_{nonlin}$.  The $(d \cdot k \times 1)$-dimension linear part is given by
\begin{equation}
    \mathbf{O}_{lin,n} = [\textbf{x}_n^T,\textbf{x}_{n-1}^T,\ldots,\textbf{x}_k^T]^T, \label{eq:Olin}
\end{equation}
and hence includes $(k-1)$ time-delay pre-images of the current state of the system.

Choosing the nonlinear part requires some skill based on insights gained from the vector field if it is known.  If there is no knowledge about the system, using a radial basis function,\cite{Scholkopf2002} for example, might be appropriate.  However, we find that a low-order polynomial function often works well even when the vector field has more complex functions.\cite{Gauthier2021}  Here, Eqs.~\ref{eq:x}-\ref{eq:u} only contain constant, linear, and quadratic terms and this motivates taking 
\begin{equation}
    \mathbf{O}_{nonlin,n} = [\mathbf{O}_{lin,n} \lceil \bigotimes \rceil \mathbf{O}_{lin,n}]^T.
\end{equation}
The operator $\lceil \bigotimes \rceil$ performs the outer product between the two vectors, selects the unique quadratic monomials, flattens the data structure, and has dimension $(d \cdot k)(d \cdot k+1)/2$.  The total feature vector is then given by
\begin{equation}
    \mathbf{O}_{total,n} = [c,\mathbf{O}_{lin,n}^T,\mathbf{O}_{nonlin,n}^T]^T,
\end{equation}
which has dimension $1+(d \cdot k)+(d \cdot k)(d \cdot k+1)/2$.

Because the NG-RC has memory due to the time-delay terms, prediction using Eqs.~\ref{eq:flow} and \ref{eq:flowNG-RC} requires data from $(k-1)$ past steps.  Thus, some data must be sacrificed to `warm up' the model.  For the NG-RC, this is only $(k-1)$ data points, whereas it is typically 10$^3$-10$^5$ data points for a traditional RC ($10^3$ was used in Ref.~\cite{Rohm2021}).  As seen below, this is important for accurate learning the co-existing attractors of Eqs.~\ref{eq:x}-\ref{eq:u} and for reproducing the basin of attraction as discussed in the next section.

\subsection{The forecasting task}

Our goal is to learn the one-step-ahead mapping of the dynamical system given by Eq.~\ref{eq:flow}.  This involves two phases:  1) a training phase based on supervised learning using time-series data produced by the system; and 2) a forecasting phase where the output of the learned model is fed back to its input, thus realizing an autonomous system that should reproduce the characteristics of the learned system.

\textbf{Training phase:} Training the NG-RC model entails collecting a block of $N_{train}$ contiguous data points, which is used to create a matrix $\textbf{Y}$ containing the difference of the state of the system at each step $\textbf{Y}_{n+1}=\textbf{x}_{n+1}-\textbf{x}_n$ over this time.  This definition is motivated by Eq.~\ref{eq:formalflow}, where we seek to learn the integral on the right hand side of the equation.  We combine this block of data with a corresponding feature vector block $\mathbf{O}_{total}$. The trained output weight matrix is obtained using regularized regression 
\begin{equation}
    \mathbb{W}_{out} = \textbf{Y} \mathbf{O}_{total}^T \left(   \mathbf{O}_{total} \mathbf{O}_{total}^T + \alpha \mathbb{I} \right)^{-1}, \label{eq:regression}
\end{equation}
where $\alpha$ is the ridge regression parameter that prevents overfitting and $\mathbb{I}$ is the identity matrix.

\textbf{Forecasting phase:} After training, the model of the learned system is given by Eqs.~\ref{eq:flow} and \ref{eq:flowNG-RC} and no external data is fed into the model.  From Eq.~\ref{eq:flowNG-RC}, we see that we require $(k-1)$ past states of $\mathbf{x}$ and thus we need to provide $k$ initial conditions of $\mathbf{x}$.  These data can be provided by a few experimental observations or evaluations of Eqs.~\ref{eq:diffeq}.  Often, this data is already available if forecasting commences immediately after training.  A traditional RC faces a similar requirement in that the reservoir needs to be `warmed up' before it can begin to make predictions and this warm up time can be substantial as mentioned above.

\subsection{Comparing actual and forecasted attractors}

Following our previous work,\cite{Rohm2021} we use a set of metrics for determining the difference between the actual and forecasted attractors so we can make a direct comparison between the performance of the NG-RC and an implementation of a traditional RC.  For both attractors, we find the time average for each variable $\langle v \rangle$ and the time average of the absolute value of the variables $\langle |v| \rangle$ with $(v=x,y,z,u)$ evaluated over 5,000 time units.

The differences between the average (absolute) forecasted $\nu$ and the ground-truth $\tilde{v}$, normalized by the absolute ground-truth averages are given by
\begin{eqnarray}
    \Delta_v \equiv \frac{\langle v \rangle - \langle \tilde{v} \rangle}{\langle |\tilde{v}| \rangle}, \\
     \Delta_{|v|} \equiv \frac{\langle |v| \rangle - \langle |\tilde{v}| \rangle}{\langle |\tilde{v}| \rangle}.
\end{eqnarray}
Here, $\Delta_v$ is a measure of the displacement of the center of mass of the attractors along each axis, while $\Delta_{|v|}$ is a measure of the difference of the extent of the attractors along each axis.  For the attractor $j$ with $j$ = (t = torus, c- = chaotic attractor with $u<0$, c+ = chaotic attractor with $u>0$), we define the metric
\begin{equation}
    \Delta_{att,j} \equiv \sqrt{\sum_i \left( \Delta_i^2 + \Delta_{|i|}^2 \right)}.
\end{equation}
To quantify the overall success of the NG-RC model for forecasting over all three attractors, we define a total error metric given by
\begin{equation}
    \Delta_{tot} \equiv \sqrt{\sum_j \Delta_{att,j}^2}.
\end{equation}

These metrics only give a sense of the forecasting accuracy of the NG-RC model but allow us to directly compare to our results reported in R\"{o}hm \textit{et al.}~\cite{Rohm2021}

\section{Results}\label{sec:results}

In this study, we learn the dynamics of the torus solution to Eqs.~\ref{eq:diffeq} and then use this learned model to forecast the torus beyond the training data set to demonstrate generalization of the model.  We then use the same model to predict the two symmetric chaotic attractors by changing the initial conditions.  This approach is similar to that taken previously in R\"{o}hm \textit{et al.},~\cite{Rohm2021} where training of the traditional RC used data from one of the chaotic attractors and the torus and symmetric attractors were forecasted.  Learning on the torus as done here is more challenging because a much smaller region of phase space is explored by the trajectory during training and hence it is believed that the model will be less accurate. Nevertheless, we show accurate prediction for all attractors.

Finally, we use the trained model to predict a section of the basin of attraction and show that it is also reproduced with high accuracy.

\subsection{Training on the torus}

\begin{figure}[th]
\includegraphics[width=\linewidth]{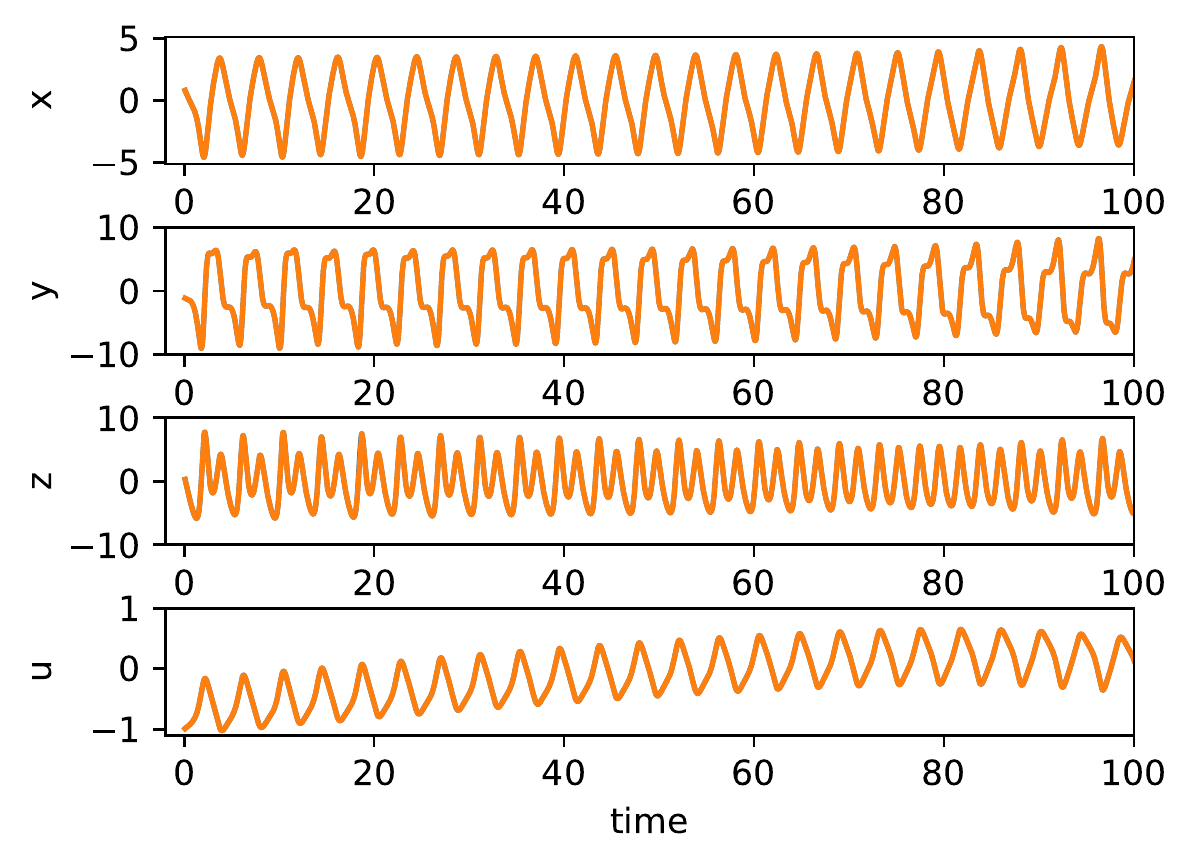}
\caption{\label{fig:torus_fine} Transient dynamics of the system with initial conditions [1,-1,1,-1]. The ground-truth data is shown in blue and the NG-RC model prediction is in orange; they overlay accurately and hence the blue trace is not visible. }
\end{figure}

We start by displaying the temporal evolution of the variables of Eqs.~\ref{eq:x}-\ref{eq:u}. The large difference in the model parameters $a$ and $b$ results in a large separation of timescales.  Figure~\ref{fig:torus_fine} shows the initial transient of the dynamics on a fine scale for initial conditions that generate the torus.  The dynamical system exhibits fast oscillations, with the fastest behavior in the $z$-variable on a time scale of $\sim 2.2$ time units.  In addition, we find also a much slower timescale in the  variation of the waveform envelope, being most apparent for the $u$-variable.

\begin{figure}[th]
\includegraphics[width=\linewidth]{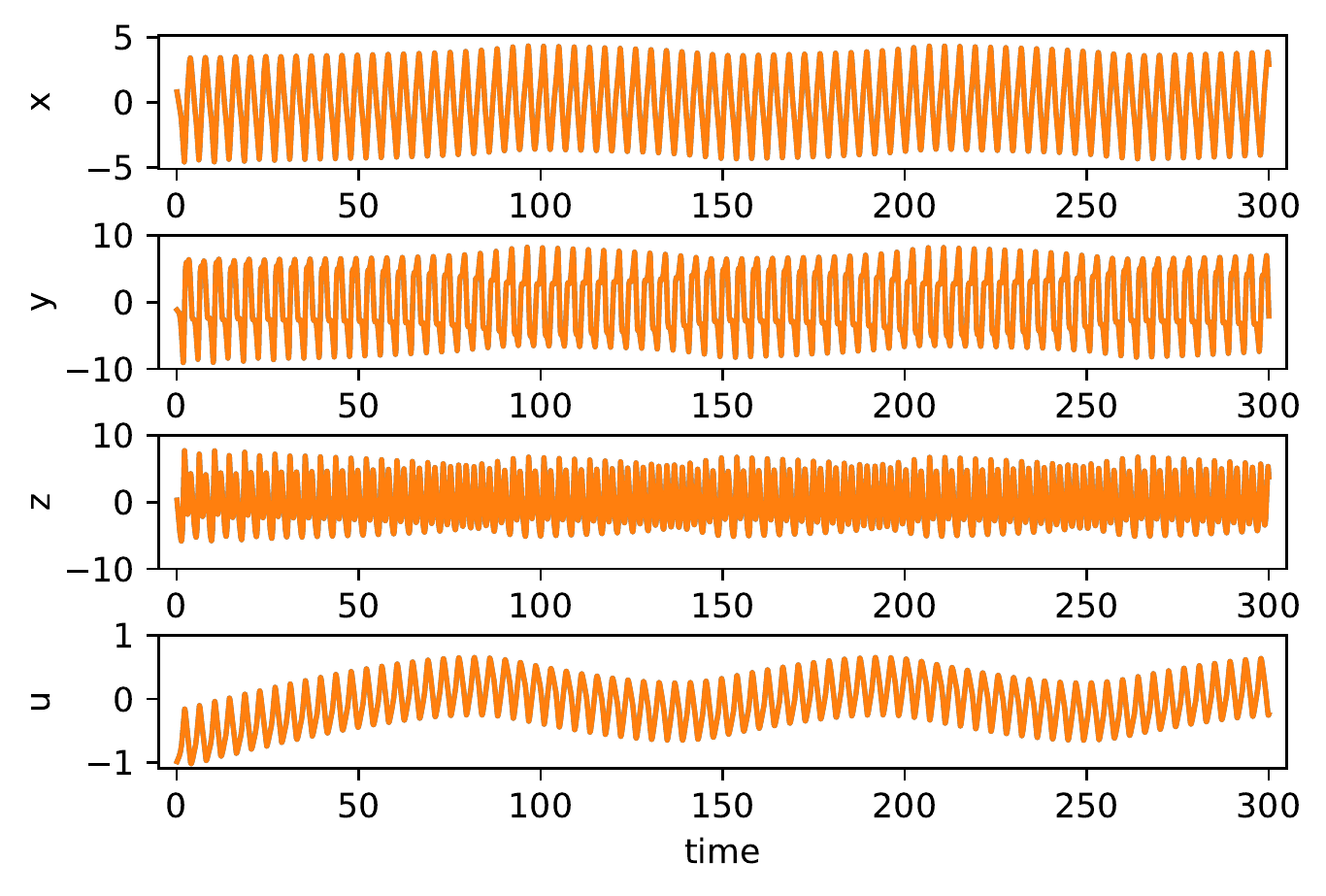}
\caption{\label{fig:torus_coarse} Transient dynamics of the system on a longer timescale with initial conditions [1,-1,1,-1]. The ground-truth data is shown in blue and the NG-RC model prediction is in orange; they overlay accurately and hence the blue trace is not visible.}
\end{figure}

Figure~\ref{fig:torus_coarse} shows the transient dynamics on a longer timescale.  From the envelope of the $u$-variable, we see that the period of the slow oscillation is $\sim$150 time units, which is $\sim$68$\times$ slower than the fastest time scale.  The slower dynamics are also apparent in the envelopes of the other variables but has a smaller effect in comparison to the $z$-variable.

In the RC literature using continuous dynamical systems as reservoirs, it has been observed that the best prediction ability is obtained when adjusting the reservoir fading memory timescale to be similar to the learned dynamics.\cite{Canaday2018}  Given the large difference in timescale of the this system, it is not obvious how to best design the reservoir: should the reservoir have nodes with a range of timescales covering both seen in Fig.~\ref{fig:torus_coarse}?  Or should there be a single node timescale to reduce the number of metaparameters that must be optimized as used in R\"{o}hm \textit{et al.}~\cite{Rohm2021}? 

In the NG-RC approach, the situation is clearer.  Our goal is to learn the flow of the dynamical system and this flow encodes both the slow and fast timescales.  We select $dt$ to be shorter than the fastest time scale (\textit{e.g.}, the sharper spike-like behavior seen in the $y$-variable of Fig.~\ref{fig:torus_fine}), but we need to train the model over a sufficiently long time period so that the NG-RC experiences the slower dynamics.  Here, we take $dt=0.05$ and the training time equal to 300 time units ($N_{train}=6,000$).  The data shown in  Fig.~\ref{fig:torus_coarse}, including the transient, is used to train the NG-RC model.

Two of the few remaining metaparameters are the values of $k$ and $\alpha$.  The value of $k$ depends on the step size $dt$.  For small $dt$ (much less than the characteristic time scale of the fastest temporal feature), $k=1$ typically works well, which can be motivated by considering the Euler method for integrating differential equations; here, only the vector field appears in the integration method, which corresponds to taking $k=1$.  Increasing $k$ only marginally improves the NG-RC testing performance.  For moderate $dt$, as used here, $k=1$ typically exhibits higher error and there can be substantial improvement using $k=2$.  This is analogous to reducing the error in numerical integrators by using data from multiple time steps. Going to $k=3$ or larger gives only modest improvement while the increase in feature vector size has a computational penalty.

For the system considered here, we explore values of $k$ up to three and find that $k=2$ gives good results for both predicting the torus as well as the chaotic attractors. For each value of $k$, we use a coarse grid search to identify the optimized value of $\alpha$; the performance is good over a wide range of the ridge regression parameter and hence the search proceeds quickly.  When taking $k=3$, the testing performance on the torus is somewhat better, but there is little improvement in the accuracy of predicting the chaotic attractors (see below).  Hence, we use $k=2$ in our studies for which $\alpha = 4.\times 10^{-5}$ is the optimal value.

The agreement between the ground-truth data generated by Eqs.~\ref{eq:diffeq} and the NG-RC model prediction during the training phase is excellent.  These data are overlayed in the figures and their difference can't be observed on this scale.  We find that the normalized root-mean-square error (NRMSE) training error is $1.5 \times 10^{-5}$.

Regarding the training computational time using the computer specified in Sec.~\ref{Sec:Methods}, we find that creating the feature vector is the most time consuming operation, taking $\sim$145 ms, whereas finding $\mathbb{W}_{out}$ only takes $\sim$2 ms.  Speeding up feature vector creation might be achieved by using the polynomial regression methods available in Scikit learn.

\subsection{Testing on the torus}

\begin{figure}[th]
\includegraphics[width=\linewidth]{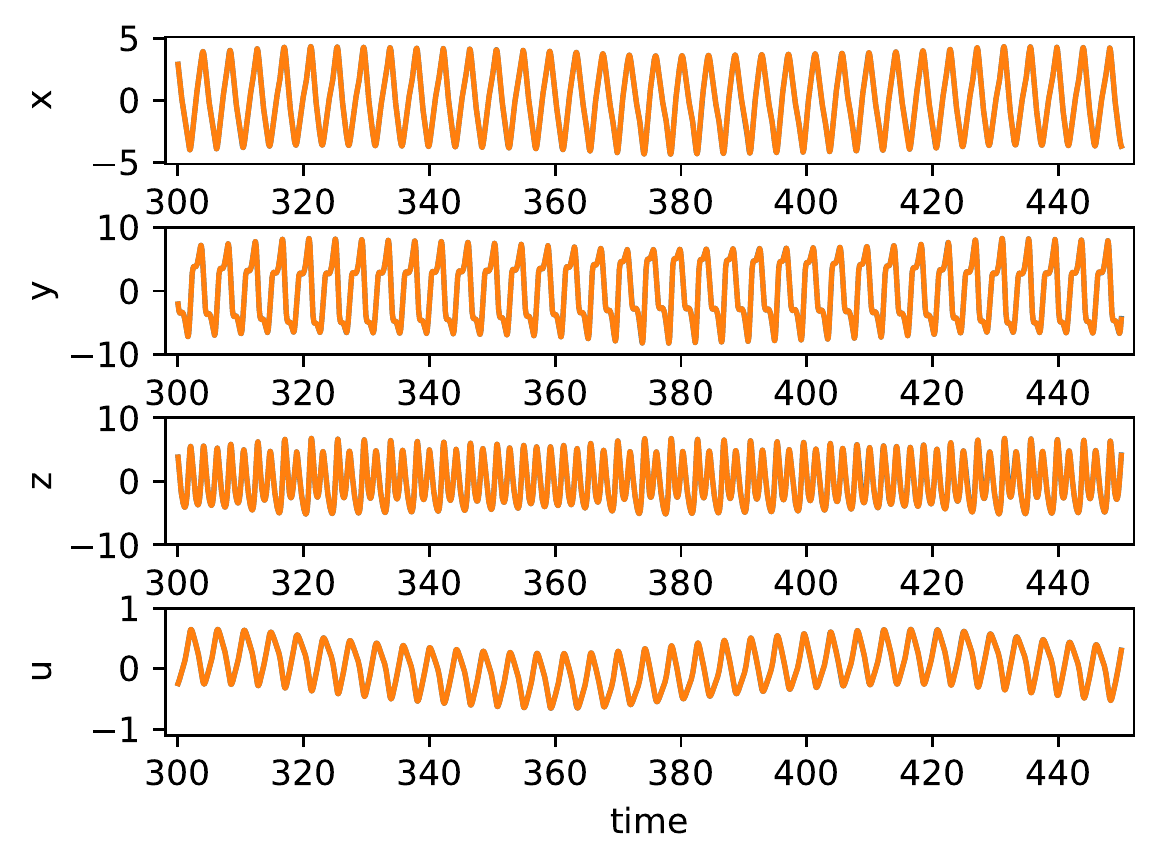}
\caption{\label{fig:torus_test} Testing the NG-RC model on the torus dynamics.  Ground truth data is shown in blue and the model prediction is in orange; they overlay accurately and hence the blue trace is not visible. The initial condition for the model is taken from the last $k=2$ data points from the training data set.}
\end{figure}

To test the generalization of the trained NG-RC model, we use it to forecast the dynamics for another 150 time units beyond the training time, which is behavior not seen during training. Figure~\ref{fig:torus_test} shows the ground-truth data and NG-RC predictions, which overlay entirely on this scale.  The NRMSE for this data is $3.3\times 10^{-3}$, indicating high-quality generalization.

The error metrics for the torus testing phase (using 5,000 time units testing time) are given in Table~\ref{tab:torus}.  We see that the largest error is in the $u$-variable where $\Delta_u \sim 1\%$. The errors for the other variables are significantly smaller.  For the torus attractor, we find that $\Delta_{att,t} = 1.2\times 10^{-2}$, largely dominated by $\Delta_u$.  For context, in R\"{o}hm \textit{et al.}~\cite{Rohm2021} we found $\Delta_u \sim 0.7$ for the torus, over a factor of $58\times$ larger than we obtain here, although in Ref.~\cite{Rohm2021}, we trained on one of the chaotic attractors rather than training on the torus as we do here.

In the next two sub-sections, we used $\mathbb{W}_{out}$ found from training on the torus, which obeys the symmetry of the system. Then we use two initial conditions that are obtained by the symmetry operation within the system.

\begin{table}[h]
\caption{\label{tab:torus} Error metrics for the torus forecasting phase. }
\begin{ruledtabular}
\begin{tabular}{lcr}
$v$&$\Delta_v$&$\Delta_{|v|}$\\
\hline \\
$x$ & $7.1\times 10^{-4}$ & $9.9\times 10^{-5}$ \\
$y$ & $3.2\times 10^{-4}$ & $-6.7 \times 10^{-5}$\\
$z$ & $-5.6\times 10^{-5}$ & $-4.7\times 10^{-4}$\\
$u$ & $1.1\times10^{-2}$ & $-5.4\times10^{-3}$ \\
\end{tabular}
\end{ruledtabular}
\end{table}

\subsection{Testing on the chaotic attractor with $u<0$}

We now use the identical trained NG-RC model but for initial conditions expected to converge towards the chaotic attractor with the average value of $u<0$, specifically $[0.,4.,0.,-5.]$.  This choice of initial condition is also used for generating the basin of attraction discussed in Sec.~\ref{Sec:basin} below following the procedure used in Ref.~\cite{Li2014}.  We need to provide the initial condition and $(k-1)=1$ additional points to initialize the model.  We generate the one required additional point using the ground-truth model (Eqs.~\ref{eq:diffeq}), but then use the learned autonomous model thereafter.

\begin{figure}[th]
\includegraphics[width=\linewidth]{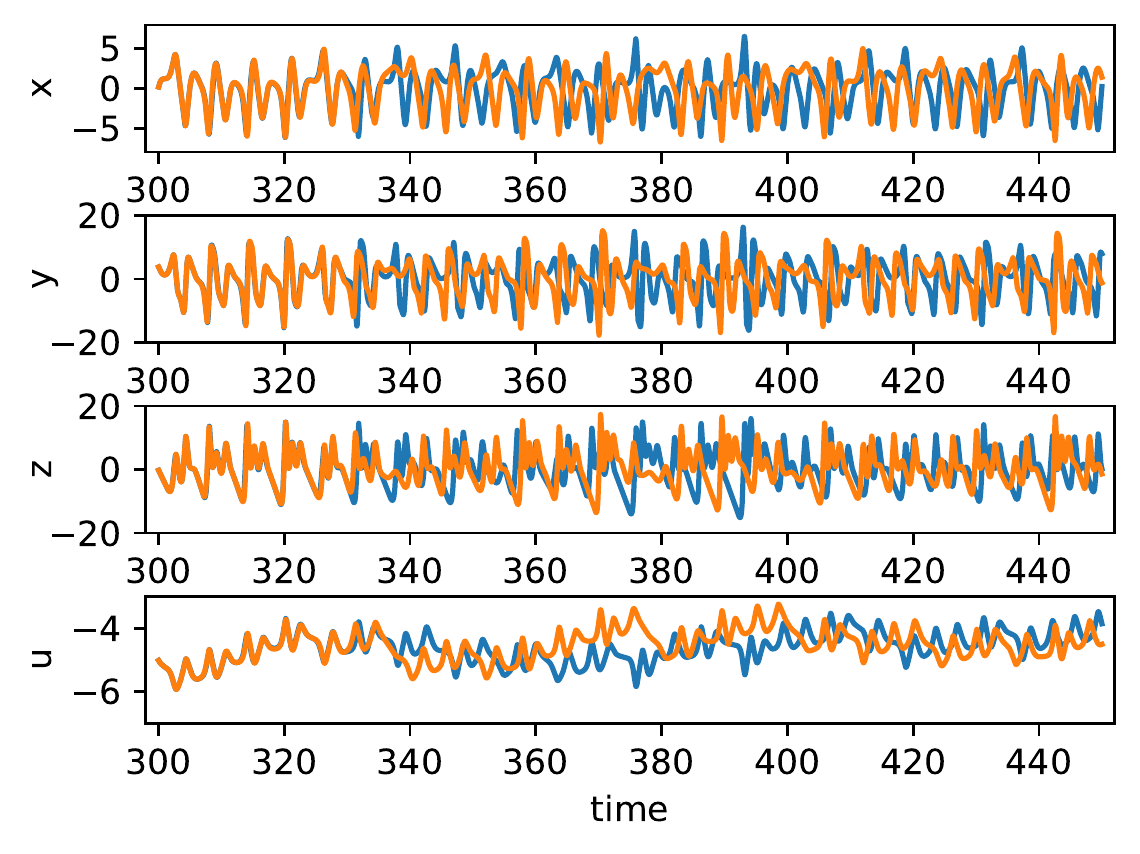}
\caption{\label{fig:chaosn} Testing the NG-RC model on the chaotic dynamics where $u<0$. The initial condition is [0.,4.,0.,-5.]. Ground truth data is shown in blue and the model prediction is in orange. }
\end{figure}

Figure~\ref{fig:chaosn} compares ground truth with NG-RC model prediction. We see that the trajectories well overlap for $\sim$30 time units (from 300 - 330 time units), which is $\sim$7.6 Lyapunov times.  The trajectories necessarily diverge because of tiny model errors and the chaotic nature of the dynamics.  Forecasting for such a long time is near the state-of-the-art for machine learning algorithms and especially stands out given the large separation of time scales in the problem, which has not been explored previously to any great extent because of the perceived hardness of this problem.

Beyond 330 time units, the forecasted dynamics is consistent with the chaotic attractor.  This is quantified using the error metrics given in Table~\ref{tab:chaosn}.  We see that all errors are below 0.4\% and the error in the $u$-variable metrics are smaller than for the torus.  The overall error metric for this attractor is $\Delta_{att,c-}=8.2\times10^{-3}$.

\begin{table}[h]
\caption{\label{tab:chaosn} Error metrics for the chaos forecasting phase with $u<0$. }
\begin{ruledtabular}
\begin{tabular}{lcr}
$v$&$\Delta_v$&$\Delta_{|v|}$\\
\hline \\ 
$x$ & $3.9\times 10^{-4}$ & $-1.1\times 10^{-3}$ \\
$y$ & $-1.9\times 10^{-4}$ & $-1.2 \times 10^{-3}$\\
$z$ & $1.1\times 10^{-3}$ & $6.1\times 10^{-3}$\\
$u$ & $3.7\times10^{-3}$ & $-3.7\times10^{-3}$ \\
\end{tabular}
\end{ruledtabular}
\end{table}

\subsection{Testing on the chaotic attractor with $u>0$}

\begin{figure}[th]
\includegraphics[width=\linewidth]{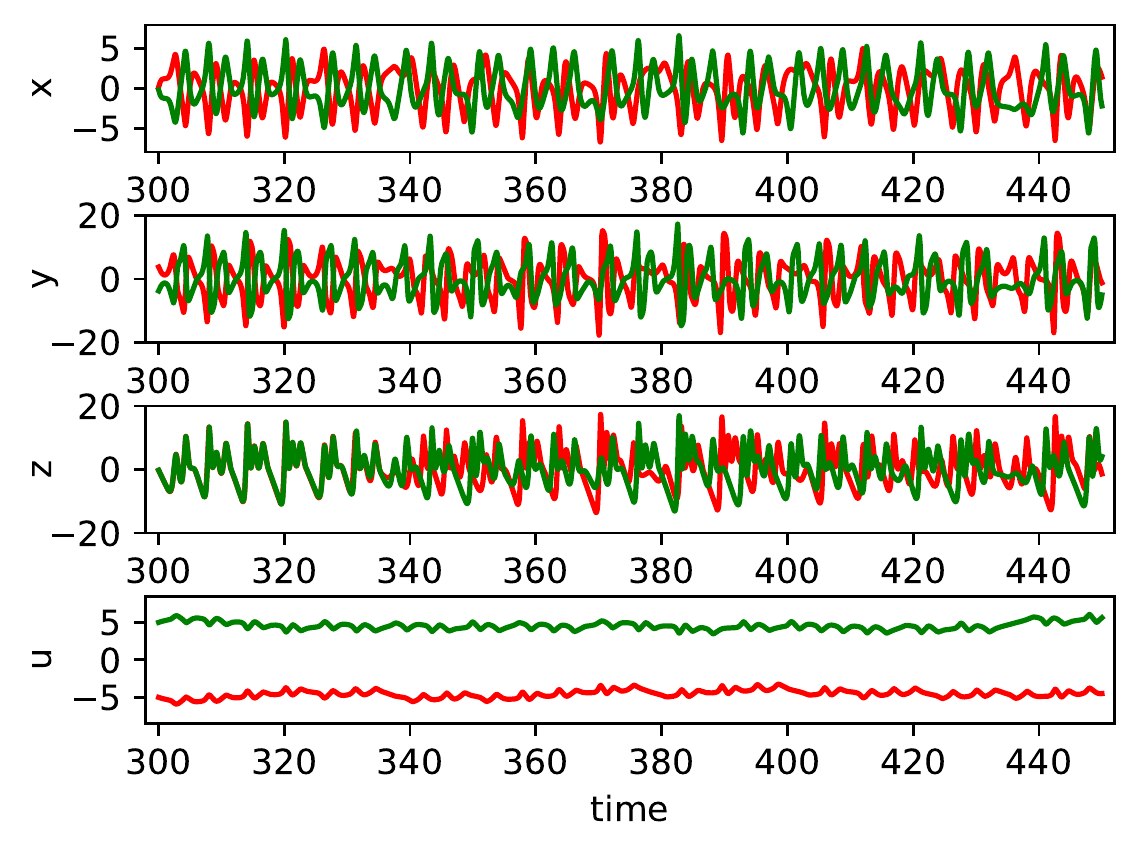}
\caption{\label{fig:chaosp} Testing the NG-RC model on the chaotic dynamics for both chaotic attractors.  Using initial condition [0.,-4.,0.,5.], the trajectory goes to the strange attractor with $u>0$ (green lines), whereas it goes to the attractor with $u<0$ (red lines) using the initial condition [0.,4.,0.,-5.].  }
\end{figure}

We also forecast the second symmetric chaotic attractor by taking the initial condition [0.,-4.,0.,5.].  Figure~\ref{fig:chaosp} shows the temporal evolution of NG-RC-forecasted behavior of both chaotic attractors.  It is seen that they initially agree with each other for $\sim$30 time units, but with the expected symmetry relations.  We do not enforce this symmetry in the NG-RC feature vector, but we see that the NG-RC model properly learns the symmetry.\cite{Barbosa2022,Barbosa2021}

Beyond 330 time units, the trajectories diverge because of the chaotic nature of the attractors, but the dynamics appears qualitatively to be part of the same attractor, which is verified by the error metrics given in Table~\ref{tab:chaosp}.  Here, the error in the $u$-variable is somewhat larger than for the other chaotic attractor but is still only $\sim$ 1\%.  The overall attractor error metric is $\Delta_{att,c+}=1.9\times 10^{-2}$ dominated by the error in the $u$-variable.  The differences in the error metrics appearing in Tables~\ref{tab:chaosn} and \ref{tab:chaosn} are largely due to the length of the testing data set size.

\begin{table}[h]
\caption{\label{tab:chaosp} Error metrics for the chaos forecasting phase with $u>0$. }
\begin{ruledtabular}
\begin{tabular}{lcr}
$v$&$\Delta_v$&$\Delta_{|v|}$\\
\hline \\ 
$x$ & $2.7\times 10^{-4}$ & $4.3\times 10^{-4}$ \\
$y$ & $2.5\times 10^{-4}$ & $-4.0 \times 10^{-3}$\\
$z$ & $3.9\times 10^{-3}$ & $-7.7\times 10^{-3}$\\
$u$ & $-1.1\times10^{-2}$ & $-1.1\times10^{-2}$ \\
\end{tabular}
\end{ruledtabular}
\end{table}

\subsection{Forecasting the basin of attraction}\label{Sec:basin}

Given the high prediction quality found in the sections above, we are motivated to see how well the trained NG-RC model predicts the basin of attraction.  The basin is complex with fractal structure\cite{Li2014} and potentially even riddled structure\cite{Sommerer1993}, where only a small change in initial conditions drives the system to different attractors.  We find the basin by selecting the initial conditions [$x_0$,$y_0$,$z_0$,$u_0$], augmenting the NG-RC model with one additional data point at time $dt$ generated by Eqs.~\ref{eq:diffeq}, and integrating for 200 time units.  If $u<-2$ ($u>2$) at this time, then the point is associated with the chaotic attractor with $u<0$ ($u>0)$, or with the torus otherwise.  We use the same procedure both with the ground-truth model and the trained NG-RC model.

\begin{figure}[th]
\includegraphics[width=\linewidth]{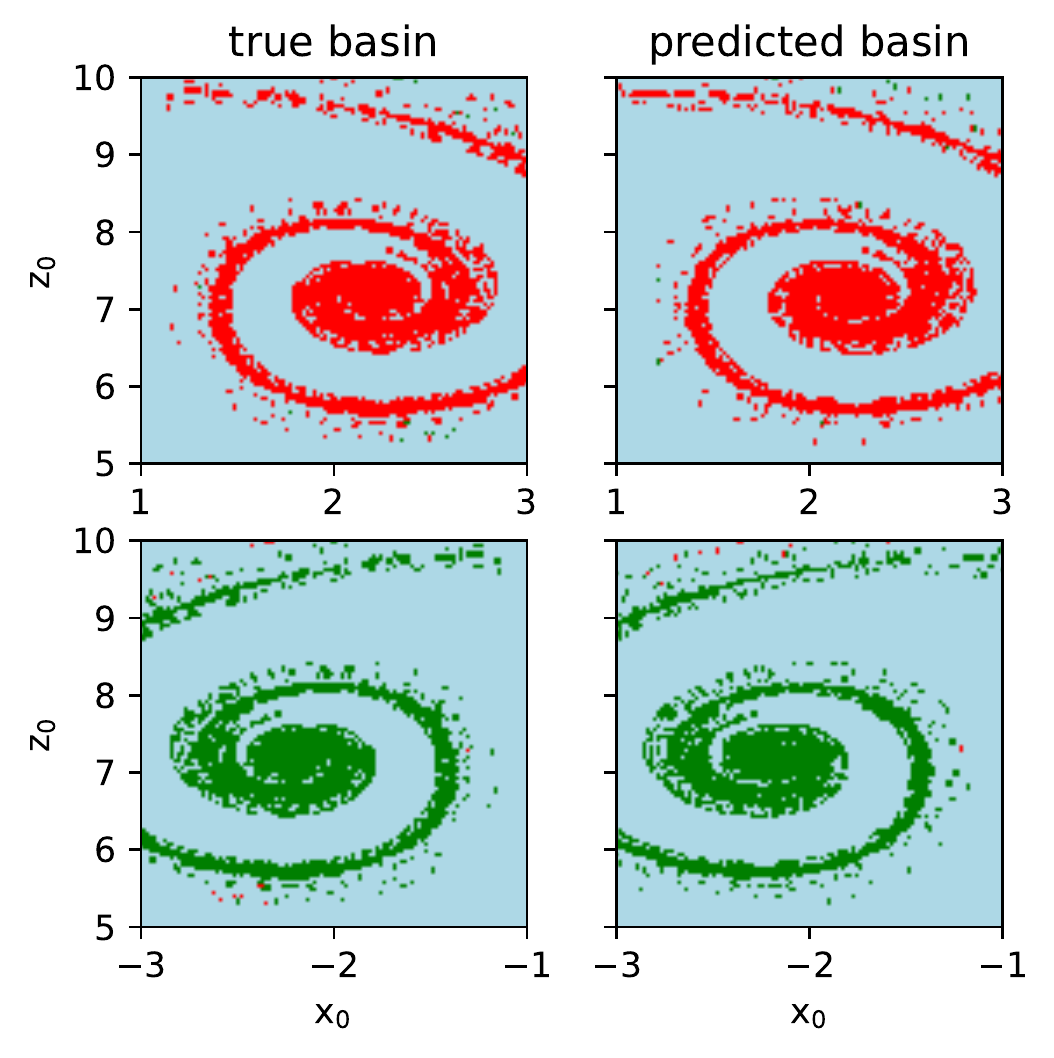}
\caption{\label{fig:basin} The basin of attraction from the prediction of Eqs.~\ref{eq:diffeq} and the trained NG-RC model evaluated on a (100 $\times$ 100) grid. The basin for the $u>0$ ($u<0$) chaotic attractor is shown in red (green) while the torus basin of attraction is shown as pale blue following the color scheme of Li and Sprott.\cite{Li2014} }
\end{figure}

We zoom in on a small region of the basin of attraction shown in Fig.~11 of Li and Sprott\cite{Li2014} to highlight its fractal structure.  Figure~\Ref{fig:basin} shows the ground-truth basins on the left and the NG-RC-forecasted basins on the right for two different regions that highlight the basin symmetry.  As mentioned above, we stress that we do not force this symmetry in the NG-RC feature vector; it learns it naturally from the training data. 

To quantify the accuracy of the NG-RC model reproducing the basins of attraction, we determine the fraction of pixels ($10^4$ total pixels) painted correctly for the two chaotic attractors.  For the chaotic attractor with $u<0$ ($u>0)$, we find that 90.0\% (90.2\%) of the pixels are painted correctly, indicating high-accuracy basin prediction.

Our ability to generate the basin of attracting is enabled by the NG-RC having a short warm up time: we only require an additional $(k-1)$ points generated by the ground truth dynamics for the data shown in Fig.~\ref{fig:basin}.  In our previous work\cite{Rohm2021} using a traditional RC, 1,000 data points from Eq.~\ref{eq:diffeq} were used to warm up the reservoir, corresponding to 200 time units (step size of 0.2 time units). The system will have already approached the final attractor in this time so the traditional RC model we used previously cannot be used to determine the basin of attraction.  Long warm-up times are present in all recurrent neural networks except for the NG-RC, which appears to use the minimum memory time possible and still achieve high prediction accuracy.

Next, we show that even the warm up points in the NG-RC can be removed using a bootstrapping method.  For simplicity, we assume $k=2$ as used throughout this work.  We generate a new NG-RC model with $k=1$ (no memory in the model) and use it to generate the one required time series data point beyond the initial condition [$x_0$,$y_0$,$z_0$,$u_0$]. This point and the initial condition are the only data needed to evaluate the $k=2$ NG-RC model.  We anticipate that the bootstrapping approach will be less accurate because the $k=1$ NG-RC model is not as accurate as the $k=2$ NG-RC model.

\begin{figure}[th]
\includegraphics[width=\linewidth]{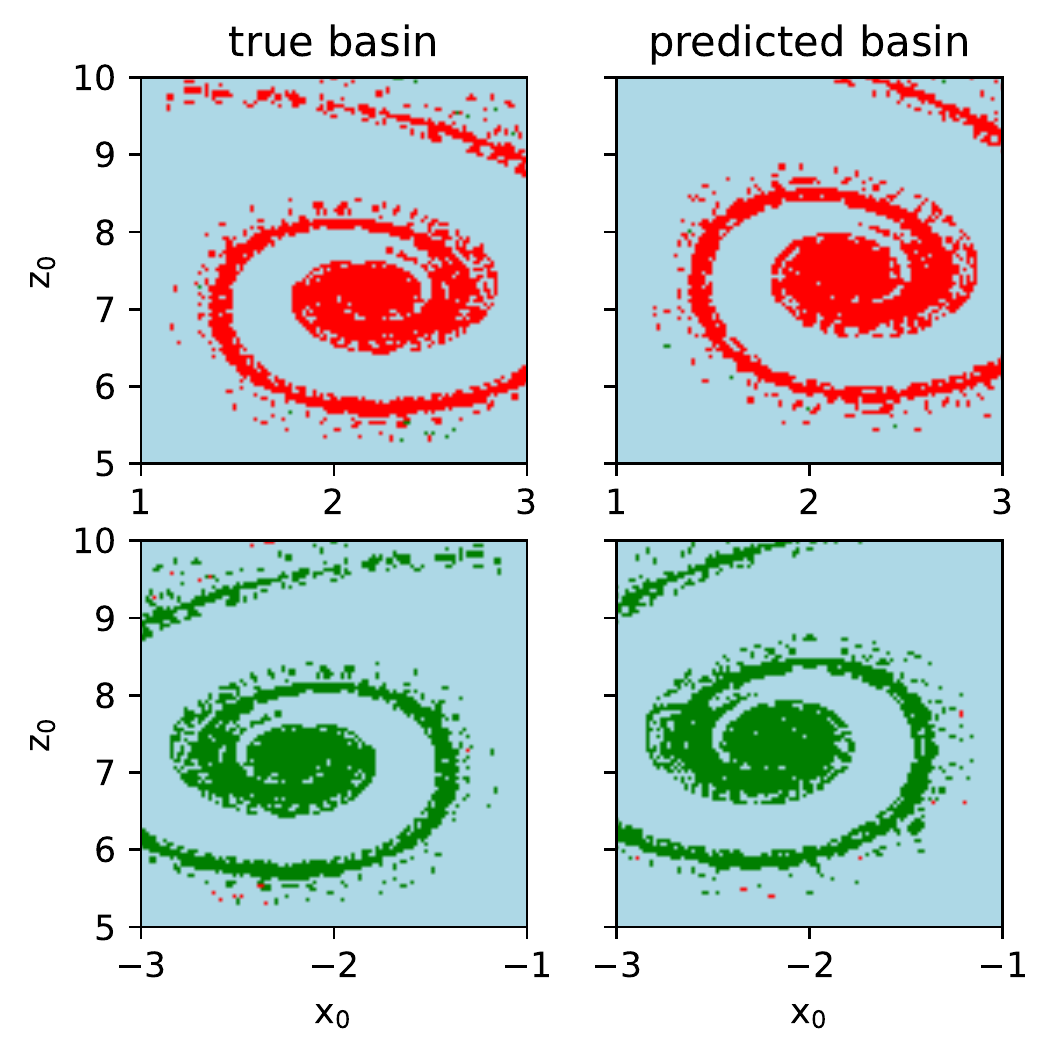}
\caption{\label{fig:basin_bootstrap} The basin of attraction from the prediction of Eqs.~\ref{eq:diffeq} and the trained NG-RC model evaluated on a (100 $\times$ 100) grid using the bootstrapping method. The color scheme is that same as in Fig.~\ref{fig:basin}. }
\end{figure}

Figure~\ref{fig:basin_bootstrap} shows the ground-truth and predicted basin of attraction using the bootstrapping method.  Qualitatively, the attractor basins appear similar with higher error in the top of the plots evident, but they continue to display the fractal or riddled structure.  For the chaotic attractor with $u<0$ ($u>0)$, we find that 80.8\% (79.8\%) of the pixels are painted correctly, indicating continued high-accuracy basin prediction. The slightly lower accuracy represents the cost of not using any ground-truth data to warm up the NG-RC.  Clearly, the bootstrapping method can be generalized for any $k$ using $(k-1)$ auxiliary NG-RC models for generating the warm up data. 

\section{Discussion}\label{sec:discussion}

We emphasize in this work that a reservoir computer trained on discretely-sampled data learns the underlying flow.  For a smooth and simple vector field such as that in Eqs.~\ref{eq:diffeq}, we expect that the flow is also smooth.  Thus, learning the flow in one region of phase space allows us to predict dynamics in other regions of phase space\cite{Hara2022} as long as the learned model is accurate.

We expect all machine learning algorithms to fail to extrapolate to other regions of phase space if the vector field is discontinuous, such as in a system with a border collision\cite{Berger2007} or other non-analytic vector fields.  We also expect that the extrapolation will be more difficult for an analytic continuous flow if the associated vector field has high-order polynomials, exponential functions, or bump functions that dominate in some regions of phase space different from where learning is performed.  Likely, criteria for learning flows and unseen attractors will require considering analytic functions and a bound on the learning accuracy, which we will explore in the future. 

The NG-RC model is particularly simple because the vector field has only quadratic nonlinear terms and hence we expect the flow to be dominated by similar terms and it has finite memory.\cite{Gauthier2021}  Here, $\mathbb{W}_{out}$ has shape (4$\times$45), resulting in 180 trainable parameters.  In contrast, $\mathbb{W}_{out}$ has shape (4$\times$300) for the traditional RC of R\"{o}hm \textit{et al.},~\cite{Rohm2021} resulting in 1,200 trainable parameters.  Having fewer trainable parameters is associated with smaller training data set sizes.\cite{Gauthier2021}  Also, solving the regression problem (Eq.~\ref{eq:regression}) scales with the length of the training data set size and quadratically on the number of feature-vector components, thus we anticipate a speedup of $\sim$75 for this part of the training.

The error metric over all three attractors for the trained NG-RC model is $\Delta_{tot} = 2.4\times 10^{-2}$, or 2.4\%.  Using a traditional RC, R\"{o}hm \textit{et al.}~\cite{Rohm2021} finds $\Delta_{tot}=2.4$.  Thus, the trained NG-RC model is 100$\times$ more accurate.  We do not understand in detail why the traditional RC didn't perform with as low of an error.  It is worth noting that the traditional RC was modelled as a continuous-time echo state network rather than the discrete-time model considered here and that may cause some differences. Also, the ground-truth model was sampled using a sample time of $0.2$ time units, whereas we use $0.05$ here. Another issue is that the traditional RC has many more metaparameters, which takes considerable computation time to optimize and there could be a small region of parameter space that gives better accuracy and was missed.  In Ref.~\cite{Rohm2021}, we also kept the decay rate of each reservoir node fixed to reduce the number of parameters needed to be optimized; this might be a source of the lower accuracy of the traditional RC, but there are likely other factors as well.

We also find that the NG-RC model can account well for disparate time scales.  The key is to take $dt$ smaller than the fastest timescale but take the training time longer than the slowest timescale.

In the future, we will explore using the NG-RC to learn the phase-space structure of higher-dimensional systems. Studies on spatio-temporal systems are already underway.\cite{Barbosa2022}  Also, we will explore using adapted basis functions or functions that focus on a region of phase space to interpolate between discontinuous regions.

\begin{acknowledgments}
DJG gratefully acknowledges the financial support of the Air Force Office of Scientific Research, Contract \#FA9550-20-1-0177. IF and AR acknowledge the Spanish State Research Agency though the Severo Ocha and Mar\'{i}a de Maeztu Program for Centers and Units of Excellence in R\&D (MDM-2017-0711) funded by MCIN/AEI/10.13039/501100011033. AR is currently an International Research Fellow of the Japan Society for the Promotion of Science (JSPS).

\end{acknowledgments}

\section*{Data Availability Statement}

The Python programs that generate all data shown in this paper are available on GitHub (*** give permanent link once accepted for publication).

\bibliography{UnseenAttractors}

\end{document}